\documentclass[sigconf]{acmart}
\usepackage{mathrsfs,multirow,algorithm,algorithmic,color}

\AtBeginDocument{%
  \providecommand\BibTeX{{%
    \normalfont B\kern-0.5em{\scshape i\kern-0.25em b}\kern-0.8em\TeX}}}

\setcopyright{acmcopyright}

\begin{document}
\title{Memory Enhanced Embedding Learning for \\Cross-Modal Video-Text Retrieval}



\author{Rui Zhao}
\authornote{Equal contribution.}
\email{rzhao62@mail.ustc.edu.cn}
\affiliation{
    \institution{University of Science and Technology of China}
}

\author{Kecheng Zheng}
\authornotemark[1]
\email{zkcys001@mail.ustc.edu.cn}
\affiliation{
    \institution{University of Science and Technology of China}
}

\author{Zheng-Jun Zha}
\email{zhazj@mail.ustc.edu.cn}
\affiliation{
    \institution{University of Science and Technology of China}
}

\author{Hongtao Xie}
\email{htxie@ustc.edu.cn}
\affiliation{
    \institution{University of Science and Technology of China}
}

\author{Jiebo Luo}
\email{jluo@cs.rochester.edu}
\affiliation{
    \institution{University of Rochester}
}
\renewcommand{\shortauthors}{}

\begin{abstract}
	Cross-modal video-text retrieval, a challenging task in the field of vision and language, aims at retrieving corresponding instance giving sample from either modality. Existing approaches for this task all focus on how to design encoding model through a hard negative ranking loss, leaving two key problems unaddressed during this procedure. First, in the training stage, only a mini-batch of instance pairs is available in each iteration. Therefore, this kind of hard negatives is locally mined inside a mini-batch while ignoring the global negative samples among the dataset. Second, there are many text descriptions for one video and each text only describes certain local features of a video. Previous works for this task did not consider to fuse the multiply texts corresponding to a video during the training. In this paper, to solve the above two problems, we propose a novel memory enhanced embedding learning (MEEL) method for video-text retrieval. To be specific, we construct two kinds of memory banks respectively: cross-modal memory module and text center memory module. The cross-modal memory module is employed to record the instance embeddings of all the datasets for global negative mining. To avoid the fast evolving of the embedding in the memory bank during training, we utilize a momentum encoder to update the features by a moving-averaging strategy. The text center memory module is designed to record the center information of the multiple textual instances corresponding to a video, and aims at bridging these textual instances together. Extensive experimental results on two challenging benchmarks, $i.e.$, MSR-VTT and VATEX, demonstrate the effectiveness of the proposed method.
\end{abstract}

\maketitle

\section{Introduction}

\begin{figure}[ht!]
    \centering
    \includegraphics[width=8cm]{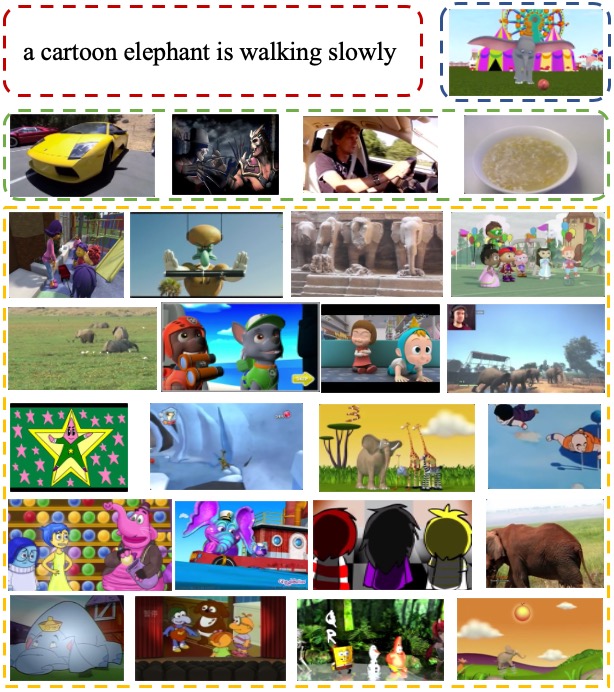}
    \caption{Illustration of a given text query, positive video sample (with {\textcolor[RGB]{36,62,124}{blue}} rectangle), negative video samples in the current mini-batch (with {\textcolor[RGB]{95,161,55}{green}} rectangle), and negative video samples in memory bank (with {\textcolor[RGB]{254,180,11}{yellow}} rectangle). A video sample is represented by a key frame for illustration. The memory bank memorizes informative and more diverse negative samples.}
    \label{Figure,illustration_data}
\end{figure}

Recent years have witnessed encouraging progress of both video analysis and natural language processing, while the cross-modal tasks 
~\cite{zha2019context,liu2018context,liu2019learning,yang2019making,zha2020adversarial,cai2014attribute,liu2019adaptive}, such as video-text retrieval~\cite{dong2018predicting,dong2019dual,song2019polysemous,li2019w2vv++,chen2020fine,mithun2018learning,liu2019use}, video captioning~\cite{gao2017video}, video moment localization~\cite{hendricks2017localizing}, and video question answering~\cite{zha2019spatiotemporal-textual} etc., remain challenging due to the semantic gap between different modalities. Cross-modal video-text retrieval~\cite{huang2018learning} aims at retrieving the target video (text) instances given text (video) query. As compared to image-text retrieval, the video-text retrieval is more challenging due to the temporal dynamics of video content and the matching across different modalities.

Existing approaches for video-text retrieval mainly focus on learning a joint feature embedding space for visual and textual modalities, where the cross-modal similarity could be measured appropriately. They generally utilize a two-branch network architecture to first encode visual and textual features respectively and then learn a joint embedding space by designed loss functions~\cite{}. The widely used loss is the bi-direction ranking loss on triplets of a video (text) embedding, a positive text (video) embedding and a negative text (video) embedding~\cite{}. It maximizes the similarity between a video (text) embedding and the corresponding text (video) embedding as well as  minimizes the similarity to all other non-matching ones. 

Although existing works have steadily improved the performance of video-text retrieval, the retrieval results are still unsatisfactory. Existing approaches update the network gradually over a series of mini-batches of training triplets. The ``receptive field" of the network is confined to the current mini-bath at each iteration. That is, the network is fed with only the current mini-batch and does not explicitly exploit the history information at previous training iterations. 
It is necessary to endow the network with memory ability towards exploiting history information and significantly enlarging its ``receptive field". Moreover, as shown in Figure~\ref{Figure,illustration_text}, a video clip is usually described by multiple text descriptions, which are different but correlated. Such kind of correlation is effective for text embedding, however not well exploited in existing works.

Motivated by the above observations, in this paper, we propose a memory enhanced embedding learning (MEEL) approach for cross-modal video-text retrieval as shown in Figure~\ref{Figure,model}. In particular, we introduce a video memory bank and a text memory bank, which memorize the video embedding features and text embedding features in a certain number of previous mini-batches, respectively. To avoid the fast evolving of feature embedding in memory banks during training, we utilize a momentum encoder to update the feature embedding in memory banks, without the requirement of gradient back-propagation. The video (text) features in memory bank are then used to help learn the text (video) feature embedding of the current mini-batch by a contrastive loss. Moreover, we use a text-center memory bank, which memorizes the

two types of memory banks, respectively: the first type is constructed for global negative mining, which includes the knowledge beyond a mini-batch, as shown in Figure~\ref{Figure,illustration_data}. There are two memory banks for this type: video memory bank and text memory bank, which record the instances' embedding before the current iteration and are updated in an enqueue-dequeue manner. While text memory bank is utilized to store more text embedding for the global negative mining of current videos mini-batch, the video memory bank is utilized to store more videos embedding for the global negative mining of the current texts mini-batch. To avoid the fast evolving of the embeddings in the memory bank during training, we utilize a momentum encoder for the embedding enqueueing and this type memory bank does not require gradient back-propagation. The second type is constructed to record the center information of the multiple instances, which includes the global knowledge across these instances and aims at bridging the difference of them during the training. There is only one memory bank for this type: text center memory bank, due to the one-to-multi property between video and text. It is trained by gradient back-propagation.
Each memory bank results in an individual loss and a memory enhanced loss is produced by the fusion of them. Both two types of memory bank can be trained respectively and either of them can boost the video-text matching performance, while fusion of them can bring further improvement because their functions are absolutely different and can learn different knowledge during the training. We conduct extensive experiments to evaluate  MEEL on two challenging datasets and achieve superior performance over the state-of-the-art approaches.

\begin{figure}[ht]
    \centering
    \includegraphics[width=8cm]{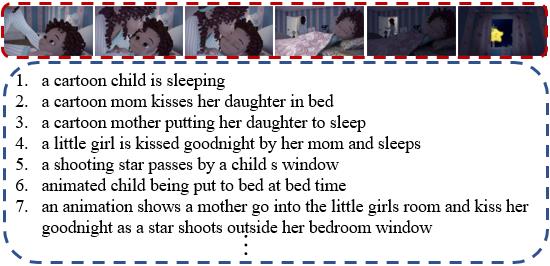}
    \caption{Illustration of the ``one video to multiple texts'' peculiarity existing in cross-modal video-text retrieval task.}
    \label{Figure,illustration_text}
\end{figure}

The main contributions of this paper are summarized as follows: 1) we introduce memory bank modules into video-text retrieval and propose novel memory enhanced embedding learning (MEEL) to incorporate complementary knowledge during training; and 2) we construct two types of memory banks: one type for global negative mining and another type for bridging the difference among the multiple texts that belong to the same video. 

\begin{figure*}[ht]
    \centering
    \includegraphics[width=\linewidth,height=9.5cm]{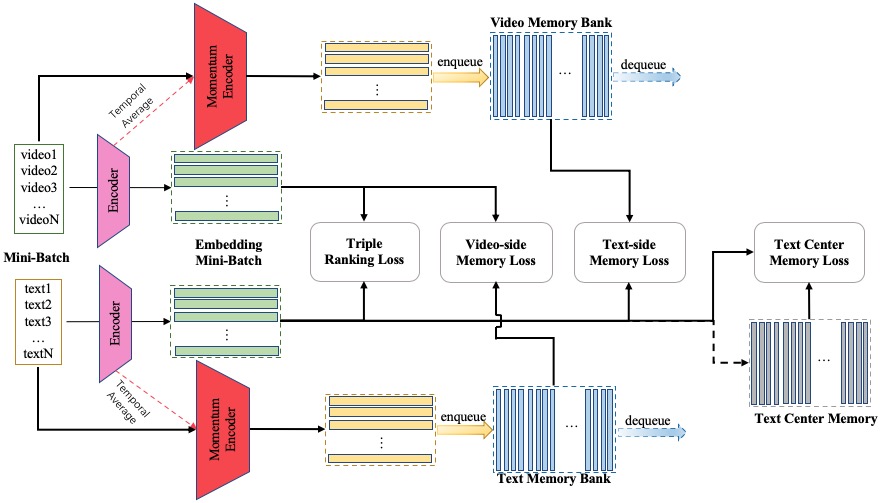}
    \caption{Illustration of our proposed architecture. It consists of two memory banks that store the embedding of videos and texts for global negative mining and one text center memory bank that aims at bridging the difference between the multiple texts of a video, leading to three individual losses that better optimize the encoder model.}
    \label{Figure,model}
\end{figure*}

\section{RELATED WORK}
\textbf{Image-Text Retrieval.} Image-text retrieval is similar to video-text retrieval, which firstly encodes images and texts into a fix-dimensional embedding vectors and aims at matching between the vectors. Most of the previous work \cite{faghri2017vse++,huang2018learning,karpathy2015deep,lee2018stacked,li2019visual} construct joint latent space learning for the matching of the embedding, which is optimized by a ranking loss that pushes the positive image-text pairs to be closer than the negative pairs. Then the distance or the similarity between any image and text pairs can be measured by cosine similarity or Euclidean distance once the joint latent space is constructed. Besides, there are also other work \cite{huang2017instance,li2017identity,wang2016learning,zhang2018deep,wang2019camp} explores to turn the matching as a classification problem, which first make a fusion between the embedding of image and text and then predict match or mismatch (+1 for match and -1 for mismatch) based on the fused multi-modal feature by a logistic regression loss optimization. \cite{lee2018stacked} proposed a stacked cross attention to make alignment between words and image regions, but only attending to either words or regions. \cite{wang2019camp} further improves it by making attending in both modalities symmetrically and exploiting message-passing between two modalities. \cite{li2019visual} propose an interpretable reasoning model to generate visual representation that captures both objects and their semantic relations.

\textbf{Video-Text Retrieval.} As mentioned before, the current dominant approaches for video-text retrieval are to learn a joint embedding space to measure the cross-modal similarity. \cite{yu2018joint} propose a joint sequence fusion model for the sequential interaction between video and text. \cite{song2019polysemous} considers the polysemous problem for videos and texts, which computes multiple and diverse representations of an instance by combining global context and locally-guided features. A recent work \cite{chen2020fine} propose a Hierarchical Graph Reasoning model, which represents complicated visual and text details into fine-grained global-local level. Hierarchical textual embeddings, which can guide the learning of diverse and hierarchical video representations, are generated by attention-based graph reasoning. We can see that all of them focus on the feature representation or the interaction of video and text. They are all trained by a simple triplet ranking loss, which ignores some key properties of video-text retrieval. Our work focus on the design of the training loss by considering two properties as mentioned before, which shows to be more effective and results in better embeddings for the matching of videos and texts.

\textbf{Memory Augmented Scheme.} Learning neural networks with memory augmented, which can provide structural and addressable knowledge for the network, has been explored in various tasks, such as question answering \cite{weston2014memory}, video understanding \cite{wu2019long}, few-shot learning \cite{santoro2016meta}, person re-identification \cite{zhong2019invariance} and so on. There are two kinds of memory: memory network and memory bank. Memory network \cite{weston2014memory} is a differentiable module and can be read and written. Memory bank is a non-parametric module and can be directly feed the feature of samples. Inspired by these work, we introduce memory bank into video-text retrieval for the complement of the two key properties. As far as we know, no study has attempted to incorporate memory bank while designing the loss for video-text retrieval.

\section{METHOD}
In this section, we firstly present the overall architecture of our proposed method as illustrated in Figure~\ref{Figure,model}, and then introduce each component in the following subsections. 

\subsection{Overall architecture}
Let $\boldsymbol{\mathcal{X}}=\{\boldsymbol{x_i},\boldsymbol{t_i}\}_{i=1}^N$ be a training set of $N$ video-text pairs.
Given a query from either modality, the goal of video-text retrieval is to identify the most relevant instances in the other modality. Previous work for this task focus on the encoder model to extract more discriminate feature representations supervised by a simple triple ranking loss. In this paper, we seek to design a more effective loss. To be specific, we introduce memory bank to augment the optimization of the network by considering two key properties: cross-modal global negative mining and one-to-multi relation between videos and texts. Just as shown in Figure \ref{Figure,model}, our architecture additionally constructs three memory banks above any off-the-shelf encoder, which consists of two cross-modal memory banks that respectively store the embedding of videos and texts for the cross-modal global negative mining and one text center memory that aims at bridging the difference between the multiple texts of a video. All the memory banks are randomly initialized. The former two cross-modal memory banks are updated by enqueuing and dequeuing: enqueue the newest embeddings of current mini-batch into the memory bank and dequeue the oldest mini-batch embeddings out of the memory bank. Besides, to avoid the fast evolving of the embeddings in the cross-modal memory bank, which may cause mismatch problem during the training, we incorporate momentum encoder for both video and text inspired by \cite{he2019momentum}. The latter text center memory are updated by gradient back-propagation.


\subsection{Cross-Modal Memory Module} \label{sec,cross-modal memory module}
Triplet ranking loss is widely adopted in many retrieval task, such as person re-identification~\cite{hermans2017defense}, image-text cross-modal retrieval~\cite{faghri2017vse++}, text-based person search~\cite{liu2019deep} and so on. Previous work~\cite{mithun2018learning,dong2019dual,liu2019use,chen2020fine} on video-text cross-modal retrieval also utilized this loss as the learning objective. It first represents a video and text into an embedding vector, and then a function $S$ is applied on them to calculate the similarity between them, which is usually the cosine similarity:
\begin{equation}
    \centering
    s_{i,j}=\frac{\boldsymbol{v}\cdot \boldsymbol{t}}{\|\boldsymbol{v}\|\cdot \|\boldsymbol{t}\|} ,
    \label{euqation,cos sim}
\end{equation}

The triplet ranking loss can then be formulated as:
\begin{equation}
\begin{aligned}
    \mathcal{L}_{tri}=max(0,\alpha-S(\boldsymbol{v},\boldsymbol{t})+S(\boldsymbol{v},\boldsymbol{t^-}))\\
    +max(0,\alpha-S(\boldsymbol{v},\boldsymbol{t})+S(\boldsymbol{v^-},\boldsymbol{t})) , \label{euqation,triplet loss}
\end{aligned}
\end{equation}
where $\boldsymbol{v}\in \mathbb{R}^d$ and $\boldsymbol{t}\in \mathbb{R}^d$ respectively denote to the final embedding of video and text with embedding dimension $d$, $\boldsymbol{t^-}$ denotes the hard negative text embedding for $\boldsymbol{v}$ and $\boldsymbol{v^-}$ denote the hard negative video embedding for $\boldsymbol{t}$ in a mini-batch.

However, just as shown in Eq.(\ref{euqation,triplet loss}), the triple ranking loss $\mathcal{L}_{tri}$ only considers the local negative samples in a mini-batch. However, there are actually many other negative samples globally existing among the training set, which is ignored by previous work. It's too time-consuming if directly comparing the samples with the whole training set at each iteration rather than in a mini-batch as mentioned in \cite{mithun2018learning}, which is not suitable for piratical application. 

While simply increase the training batch can make comparison with more samples, the optimization may be difficult and the memory of GPU can also be limited. Thus a suitable and effective strategy needs to be explored. In this paper, we introduce memory bank for the cross-modal global negative mining. We define two complimentary formulations of cross-modal memory below: \textit{Text-to-Video} and \textit{Video-to-Text}.

\textbf{Video-to-Text Cross-Modal Memory.} To mine the global negative samples for video, we construct a memory bank $\boldsymbol{M_t}\in \mathbb{R}^{K\times d}$ in text side to store the previous text embeddings before current mini-batch. Once given the video embedding $\boldsymbol{v}$ from mini-batch with corresponding ground-truth texts set $\boldsymbol{T_v}$ and the text memory bank $\boldsymbol{M_t}$, we incorporate a contrastive loss function named InfoNCE \cite{oord2018representation} to optimize the encoder model:
\begin{equation}
    \mathcal{L}_{v2t}=-log{\frac{exp(\boldsymbol{v}\cdot \boldsymbol{k_+}/\tau)}{\sum\limits_{\substack{i=0 \\ \boldsymbol{k_i}\notin \boldsymbol{T_v}}}^{K-1} exp(\boldsymbol{v}\cdot \boldsymbol{k_i}/\tau)}} ,
    \label{equation,v2t}
\end{equation}
where $\tau$ denotes a temperature hyper-parameter, $K$ denotes the length of memory bank. $\boldsymbol{k_+}$ denotes the current positive text embedding for $\boldsymbol{v}$, while $\boldsymbol{k_i}$ denotes the negative text embedding. Note that, because of the one-to-multi relation between videos and texts, there may be multiple ground-truth texts in text memory bank for current video, we implement mask operation for these texts. Thus, the sum is over one positive and other negative samples. This loss is intuitively a log loss of softmax-based classifier that aims at classifying the positive $(\boldsymbol{v},\boldsymbol{k+})$ pair against the remaining negative pairs. 

\textbf{Text-to-Video Cross-modal Memory.} Same to video-to-text cross-modal memory, we also construct a memory bank $\boldsymbol{M_v} \in \mathbb{R}^{K\times d}$ in video side to mine the global negative samples for text. Given the embedding $\boldsymbol{t}$ of a text from mini-batch with corresponding ground-truth video set $\boldsymbol{V_t}$ and the text memory bank $\boldsymbol{M_v}$, the contrastive loss between them can be formulated as:
\begin{equation}
    \mathcal{L}_{t2v}=-log{\frac{exp(\boldsymbol{t}\cdot \boldsymbol{k_+}/\tau)}{\sum\limits_{\substack{i=0\\\boldsymbol{k_i}\notin \boldsymbol{T_v}}}^{K-1} exp(\boldsymbol{t}\cdot \boldsymbol{k_i}/\tau)}} ,
    \label{equation,t2v}
\end{equation}
where $\tau$ denotes a temperature hyper-parameter. $\boldsymbol{k_+}$ denotes the current positive video embedding for $\boldsymbol{t}$, while $\boldsymbol{k_i}$ denotes the negative video embedding. Mask operation is also adopted.

\textbf{Momentum Encoder} As mentioned before, if we directly feed the embedding of current mini-batch into the memory bank, the embedding in the memory bank may fast evolve during the training. To avoid this, we incorporate a momentum encoder, as inspired by \cite{he2019momentum}. Define the original encoder model as $f_q$ and the momentum encoder as $f_k$. The model structure of $f_q$ and $f_k$ are completely same, while the only difference is the way of parameters updating. The parameters updating of $f_k$ can be formulated as:
\begin{equation}
    \theta_k\leftarrow m\theta_k+(1-m)\theta_q , \label{equation,momentum}
\end{equation}
where $m\in [0,1)$ is a momentum hyper-parameter, $\theta_k$ is the parameter of $f_k$ and $\theta_q$ is the parameter of $f_q$. Only the parameter $\theta_q$ is updated by gradient back-propagation and the momentum update in Eq.(\ref{equation,momentum}) makes the evolving of $\theta_k$  more smoothly than $\theta_q$. In this way, though the embeddings in memory bank are encoded in different mini-batch, the difference between them can be reduced by applying a large momentum. Besides, in our experiments, testing by the embeddings of $f_k$ shows a better performance than $f_q$, which verifies the effectiveness of the smoothing operation.  Algorithm~\ref{alg:1} provides the pseudo-code of the overall cross-modal memory module with momentum encoder for video-text retrieval.



\subsection{Text Center Memory Module} 
As mentioned before, due to the complexity of video content and the variance of text description, a video can usually be retrieved by several different but semantic related sentences. This kind of one-to-multi relation between video and text are necessary to be considered. In order to bridge the difference between the multiple texts belonging to a same video, we proposed to construct a text center memory $\boldsymbol{M_c}\in \mathbb{R}^{H\times d}$, which records the center information of these texts.
We regard the texts that belong to one video as one class and adopt intra-class variation minimization by:
\begin{equation}
    \mathcal{L}_c=\frac{1}{2}\sum_{i=1}^{B} \left\| \boldsymbol{t_i}-\boldsymbol{c_{y_i}} \right\|_2^2 , \label{equation,loss}
\end{equation}
where $\boldsymbol{t_i}$ denotes the text embedding from mini-batch that belongs to $y_i$th class and  $\boldsymbol{c_{y_i}}$ is corresponding center vector of $y_i$th class in $\boldsymbol{M_c}$. $B$ and $H$ are the size of mini-batch and the text center memory.

Instead of updating the centers with respect to the entire training set, we perform the update based on the mini-batch following to \cite{wen2016discriminative}. In each iteration, we compute the distance between the text embeddings of current mini-batch and corresponding class center embedding in text center memory as $\mathcal{L}_c$.

\begin{algorithm}[htbp]
\caption{Pytorch Pseudocode for Cross-Modal Memory Module}
\label{alg:1}
\begin{algorithmic}
\STATE \# f\_q, g\_q: the original encoder network of video and text
\STATE \# f\_k, g\_k: the momentum encoder network of video and text
\STATE \# queue\_v, queue\_t: embedding memory bank of video and text 
\STATE \# m: momentum coefficient
\STATE \# t: temperature coefficient
\STATE initializing: f\_k.params=f\_q.params, g\_k.params=g\_q.params
\FOR{(x\_v, x\_t) in dataloader} 
\STATE v\_q = f\_q.forward(x\_v)
\STATE t\_q = g\_q.forward(x\_t)
\STATE v\_k = f\_k.forward(x\_v).detach() {\textcolor[RGB]{128,138,135}{\# no gridient}}
\STATE t\_k = g\_k.forward(x\_t).detach()
\STATE {\textcolor[RGB]{128,138,135}{\# logits for positive: Bx1}}
\STATE l\_pos\_v2t = bmm(v\_q.view(B, 1, D), t\_k.view(B, D, 1))
\STATE l\_pos\_t2v = bmm(t\_q.view(B, 1, D), v\_k.view(B, D, 1))
\STATE {\textcolor[RGB]{128,138,135}{\# logits for negative: BxK}}
\STATE l\_neg\_v2t = mm(v\_q.view(B, D), queue\_t.view(D, K))
\STATE l\_neg\_t2v = mm(t\_q.view(B, D), queue\_v.view(D, K))
\STATE {\textcolor[RGB]{128,138,135}{\# logits: Bx(1+K)}}
\STATE logits\_v2t = concat([l\_pos\_v2t, l\_neg\_v2t], dim=1)
\STATE logits\_t2v = concat([l\_pos\_t2v, l\_neg\_t2v], dim=1)
\STATE {\textcolor[RGB]{128,138,135}{\# constractive loss in Eq.(\ref{equation,v2t}) and Eq.(\ref{equation,t2v})}}
\STATE labels = zeros(B)
\STATE loss\_v2t = CrossEntropyLoss(logits\_v2t / t, labels)
\STATE loss\_t2v = CrossEntropyLoss(logits\_t2v / t, labels)
\STATE loss=loss\_tri+loss\_v2t+loss\_t2v {\textcolor[RGB]{128,138,135}{\# loss\_tri denotes triplet loss}}
\STATE {\textcolor[RGB]{128,138,135}{\# SGD update}}
\STATE loss.backward()
\STATE Update(f\_q.params, g\_q.params)
\STATE {\textcolor[RGB]{128,138,135}{\# Momentum update}}
\STATE f\_k.params = m  * f\_k.params + (1-m) * f\_q.params
\STATE g\_k.params = m * g\_k.params + (1-m) * g\_q.params
\STATE {\textcolor[RGB]{128,138,135} {\# Memory update}}
\STATE Enqueue(queue\_v, v\_k)
\STATE Dequeue(queue\_v)
\STATE Enqueue(queue\_t, t\_k)
\STATE Dequeue(queue\_t)
\ENDFOR
\end{algorithmic}
\end{algorithm}

\subsection{Training and Testing}
The final loss function that is used to train the whole model is the summation of the triple ranking loss ($\mathcal{L}_{tri}$), video-side memory loss ($\mathcal{L}_{v2t}$), text-side memory loss ($\mathcal{L}_{t2v}$) and the text center memory loss ($\mathcal{L}_c$):
\begin{equation}
    \mathcal{L}=\mathcal{L}_{tri}+\mathcal{L}_{v2t}+\mathcal{L}_{t2v}+\alpha\mathcal{L}_{c} ,
\end{equation}
where $\alpha$ is a scale weight to balance the influence of the loss term. The additional memory can capture different external knowledge, which can improve the optimization of the encoder model to obtain a more robust feature representation with the joint supervision of $\mathcal{L}_{tri}$, $\mathcal{L}_{v2t}$, $\mathcal{L}_{t2v}$ and $\mathcal{L}_{t2v}$.

During the testing time, given a text description or a video clip, its final representation is extracted by its corresponding momentum encoder network. Assume there are $p$ videos and $q$ texts in the whole test set, the distance between all the video-text pairs are calculated by cosine similarity in Eq.(\ref{euqation,cos sim}), which produce a distance matrix $D_{cos}\in \mathbb{R}^{p\times q}$. The distance are then sorted and $Recall@1$ through $Recall@10$ are reported.

\section{EXPERIMENTS}
\begin{table*}[ht]
\small
\begin{center}
\caption{Overall performance comparison with the state-of-the-art methods on the MSR-VTT dataset. Higher $R@K$ and lower $MedR$, $MeanR$ is better. Sum of Recalls ($RSum$) indicates the overll performance.}\label{table,msrvtt}
\begin{tabular}{c|ccccc|ccccc|c} 
\hline
\hline
\multirow{2}{*}{Method} & \multicolumn{5}{c|}{Text-to-Video Retrieval} & \multicolumn{5}{c|}{Video-to-Text Retrieval} & \multirow{2}{*}{Sum of Recalls}  \\ 
\cline{2-11}
                        & R@1 & R@5  & R@10 & MedR & MeanR             & R@1  & R@5  & R@10 & MedR & MeanR            &                                  \\ 
\hline
W2VV~\cite{dong2018predicting}                    & 1.8 & 7.0  & 10.9 & 193  & -                 & 9.2  & 25.4 & 25.4 & 24   & -                & 90.3                             \\ 

VSE~\cite{kiros2014unifying}                     & 5.0 & 16.4 & 24.6 & 47   & 215.1             & 7.7  & 20.3 & 31.2 & 28   & 185.8            & 105.2                            \\

Mithun et al.~\cite{huang2018learning}           & 5.8 & 17.6 & 25.2 & 61   & 296.6             & 10.5 & 26.7 & 35.9 & 25   & 266.6            & 121.7                            \\ 

W2VViml~\cite{dong2019dual}                 & 6.1 & 18.7 & 27.5 & 45   & -                 & 11.8 & 28.9 & 39.1 & 21   & -                & 132.1                            \\ 

VSE++~\cite{faghri2017vse++}                   & 5.7 & 17.1 & 24.8 & 65   & 300.8             & 10.2 & 25.4 & 35.1 & 25   & 228.1            & 118.3                            \\ 
DualEncoding~\cite{dong2019dual}            & 7.7 & 22.0 & 31.8 & 32   & -                 & 13.0 & 30.8 & 43.3 & 15   & -                & 148.6                            \\ 
\hline
VSE++ with MEEL       & 6.2 & 18.9 & 27.5 & 43   & 224.7             & 11.0 & 28.2 & 38.8 & 20   & 150.7            & 130.6                            \\

DualEncoding with MEEL         & \textbf{8.3} & \textbf{24.1} & \textbf{34.4} & \textbf{26} & \textbf{165.6} & \textbf{15.5} & \textbf{35.4} & \textbf{46.2} & \textbf{12} & \textbf{99.4} & \textbf{163.9}                   \\
\hline
\hline
\end{tabular}
\end{center}
\end{table*}


\begin{table*}[ht]
\small
\begin{center}
\caption{Performance comparision with the state-of-the-art methods on the VATEX dataset.} \label{table,vatex}
\begin{tabular}{c|ccccc|ccccc|c} 
\hline\hline
\multirow{2}{*}{Method} & \multicolumn{5}{c|}{Text-to-Video Retrieval}                                 & \multicolumn{5}{c|}{Video-to-Text Retrieval}                                 & \multirow{2}{*}{Sum of Recalls}  \\ 
\cline{2-11}
                        & R@1           & R@5           & R@10          & MedR         & MeanR         & R@1           & R@5           & R@10          & MedR         & MeanR         &                                  \\ 
\hline
VSE~\cite{kiros2014unifying}        & 18.9          & 50.7          & 64.9          & 5          & 45.2          & 25.6          & ~58.4~        & 71.8          & 4          & 35.8          & 290.3                            \\ 

VSE++~\cite{faghri2017vse++}        & 21.1          & 53.5          & 66.2          & 5          & 50.1          & 30.2          & 61.2          & 74.8          & \textbf{3} & 33.9          & 307.0                            \\ 

DualEncoding~\cite{dong2019dual}    & 23.7          & 58.8          & 71.2          & \textbf{4} & 41.7          & 32.1~         & ~63.4~        & 75.1          & \textbf{3} & 35.0          & 324.3                            \\ 
\hline
VSE++ with MEEL                   & 22.1          & 55.5          & 68.7          & \textbf{4}          & 36.7          & 32.1          & 63.0          & 75.9          & \textbf{3} & \textbf{27.8} & 317.3                            \\

DualEncoding with MEEL            & \textbf{25.4} & \textbf{61.6} & \textbf{73.4} & \textbf{4} & \textbf{36.5} & \textbf{35.5} & \textbf{66.2} & \textbf{76.4} & \textbf{3} & 30.5 & \textbf{338.6}                   \\
\hline\hline
\end{tabular}
\end{center}
\end{table*}

\begin{table*}[]
\small
\begin{center}
\caption{Ablation studies on the MSR-VTT dataset to investigate the additions of the three memory based loss.}\label{table,msrvtt_ab}
\begin{tabular}{c|c|c|c|c|ccccc|ccccc|c}
\hline\hline
\multicolumn{1}{l|}{\multirow{2}{*}{$\mathcal{L}_{tri}$}} & \multicolumn{1}{l|}{\multirow{2}{*}{$\mathcal{L}_{v2t}$}} & \multicolumn{1}{l|}{\multirow{2}{*}{$\mathcal{L}_{t2v}$}} & \multicolumn{1}{l|}{\multirow{2}{*}{$\mathcal{L}_c$}} & \multicolumn{1}{l|}{\multirow{2}{*}{Momentum}} & \multicolumn{5}{c|}{Text-to-Video Retrieval} & \multicolumn{5}{c|}{Video-to-Text Retrieval} & \multirow{2}{*}{Sum of Recalls} \\ \cline{6-15}
\multicolumn{1}{l|}{}                         & \multicolumn{1}{l|}{}                         & \multicolumn{1}{l|}{}                         & \multicolumn{1}{l|}{}                       & \multicolumn{1}{l|}{}                          & R@1           & R@5   & R@10  & MedR & MeanR & R@1     & R@5    & R@10   & MedR   & MeanR   &                                 \\ \hline
$\surd$                                       &                                               &                                               &                                             &                                                & 7.7           & 22.0  & 31.8  & 32   & -     & 13.0    & 30.8   & 43.3   & 15     & -       & 148.6                           \\ \hline
$\surd$                                       & $\surd$                                       & $\surd$                                       &                                             &                                                & 8.1           & 23.4  & 32.8  & 32   & 206.4 & 13.9    & 32.3   & 44.5   & 15     & 120.5   & 155.0                           \\ \hline
$\surd$                                       & $\surd$                                       & $\surd$                                       &                                             & $\surd$                                        & 8.2           & 24.1  & 34.4  & \textbf{26}   & 176.3 & 14.7    & 34.1   & 45.3   & 13     & 110.7   & 159.2                           \\ \hline
$\surd$                                       &                                               &                                               & $\surd$                                     &                                                & 7.8           & 23.2  & 33.3  & 27   & 178.0 & 14.5    & 33.8   & 44.6   & 14     & 102.0   & 157.2                           \\ \hline
$\surd$                                       & $\surd$                                       & $\surd$                                       & $\surd$                                     & $\surd$                                        & \textbf{8.3}  & \textbf{24.1}  & \textbf{34.4}  & \textbf{26}   & \textbf{165.6} & \textbf{15.5}    & \textbf{35.4}   & \textbf{46.2}   & \textbf{12}     & \textbf{99.4}    & \textbf{163.9}                           \\ \hline\hline
\end{tabular}
\end{center}
\end{table*}

In this section, we conduct extensive experiments to evaluate the effectiveness of our proposed MEEL. We apply MEEL above the state-of-the-art methods on two challenging datasets (MSR-VTT~\cite{xu2016msr} and VATEX~\cite{wang2019vatex}) for video-to-text retrieval and text-to-video retrieval, which are widely used in other vision and language task such as video question answering and video captioning. Moreover, ablation study is also shown.

\subsection{Experimental Setup}
\textbf{Datasets.} In this paper, we first conduct experiments on MSR-VTT and the newly released VATEX dataset. MSR-VTT is the dominant dataset for video-text cross-modal retrieval. It consists of 10000 video clips, each of which is annotated with 20-sentence descriptions. All the sentences are utilized in our work. Following the prior work \cite{mithun2018learning, dong2019dual}, we split MSR-VTT into 6513 clips for training, 497 clips for validation and 2990 clips for testing. The VATEX datasets contains 25,991 videos for training, 3000 for validation and 6000 for testing. Since the annotations on testing set are private, we regard the original validation set as new testing set and randomly split 1500 videos from orginal training set as new validation set. The remaining 24491 videos compose the new training set. While there are 10 sentence descriptions in English and Chinese respectively describing each video, we only utilize the English part in our work. 

\textbf{Evaluation Metrics.} There are two kinds of video-text retrieval task: 1) sentence retrieval task requires identifying ground truth sentences given a query video (V2T); 2) video retrieval task requires identifying ground truth videos given a query sentences (T2V). We measure the retrieval performance for both V2T and T2V task with the standard metrics commonly used in information retrieval: Recall at $K$ ($R@K$), Median Rank ($MedR$) and Mean Rank ($MeanR$). $R@K$ is defined as the recall rate at the top $K$ results to the query, where $K$ is usually set as $1, 5, 10$, where higher score indicates better performance. $MedR$ and $MeanR$ are respectively the median and average rank of ground truth samples in the retrieved ranking list, where lower score achieves better performance. $Rsum$ is calculated by summing all the $R@K$ in V2T and T2V to compare the overall performance.

\textbf{Implementation Details.} All the implementations are based on Pytorch framework with NVIDIA 1080Ti gpu. We keep the settings of encoder model consistent with prior work. For the video encoding, Resnet-152 pretrained on ImageNet~\cite{he2016deep} is adopted to extract frame-wise video feature for MSR-VTT dataset, while the officially provided I3D~\cite{carreira2017quo} video feature is utilized for VATEX dataset. For the text encoding, we set the word embedding as 512 and initialize by word2vec~\cite{mikolov2013efficient} pretrained on English tags of 30 million Flickr images following to \cite{dong2019dual}. SGD with Adam is adopted as the optimizer. The size of our video memory bank and text memory $K$ and the temperature coefficient $\tau$ in Eq.(\ref{equation,v2t}) and Eq.(\ref{equation,t2v}) are set to 2560 and 0.07 respectively. Momentum coefficient $m$ in Eq.(\ref{equation,momentum}) is set to 0.99 and then decays to 0.999 after 2 epochs. $\alpha$ in Eq.(\ref{equation,loss}) is set to 0.005.

\subsection{Comparison with the State of the Art}
We compare our method with several recent state-of-the-art models on MSR-VTT and VATEX datasets, including W2VV~\cite{dong2016word2visualvec}, VSE~\cite{kiros2014unifying}, VSE++~\cite{faghri2017vse++}, Mithun et al.~\cite{mithun2018learning} and DualEncoding~\cite{dong2019dual}. Note that, we apply our proposed MEEL on two state-of-the-art methods DualEncoding and VSE++ to show its effectiveness. 

Table \ref{table,msrvtt} shows the overall I2T and T2I retrieval result of our method on MSR-VTT testing set. Our proposed MEEL improves the state-of-the-art model DualEncoding by a margin with respect to all the retrieval metric. Specifically, DualEncoding with MEEL achieves the best $R@1$ result: 8.3 for T2V and 15.5 for V2T, leading to 7.8\% and 19.2\% relative gains. The overall performance reflected by the $RSum$ metric is also boosted with a margin +15.3, which indicates that MEEL can optimize the encoder model better and results in a discriminate feature representation. Note that, the external memory requires only little extra GPU space and additional training time, which can be easy insert to other model.

To further demonstrate the robustness of our approach, we further provide quantitative results on VATEX dataset in Table \ref{table,vatex}. The difference is that video feature of VATEX is extracted by I3D model, officially provided by \cite{wang2019vatex}. We can see that after applying MEEL on DualEncoding, the model achieves consistent improvement across different dataset and feature, which result in a relative 7.2\% and 10.6\% gains on $R@1$ metric and +14.3 marin on the over performance $RSum$. This verifies that it's beneficial to take global negative mining and the one-to-multi-relation into consideration for video-text retrieval task.

To verify the generalization ability of the proposed MEEL, we also apply it on VSE++, which results in a consistent improvement on both datasets.

\subsection{Ablation Studies}

In order to demonstrate the effectiveness and contribution of each component of our proposed MEEL, we apply it on DualEncoding and conduct a series of ablation experiments on MSR-VTT dataset. We first evaluate the effect of the cross-modal memory module, the text center memory module, the momentum encoder and compare their performance for both video retrieval and sentence retrieval task. Then we evaluate how $RSum$ is affected by the memory size of the cross-modal memory module.

\textbf{Impact of proposed component.} Tables \ref{table,msrvtt_ab} summarizes the ablation results. We first remove our center memory module and we can observe that the video-side memory loss ($\mathcal{L}_{v2t}$) and the text-side memory loss ($\mathcal{L}_{t2v}$) together result in an improvement from 148.6 to 159.2 over $Rsum$ by a margin, which verifies that the global negative mining are necessary for video-text retrieval. We also remove our cross-modal memory module and only utilize the center memory module, which achieves an improvement on $Rum$. These two comparisons show the importance to explore external knowledge augmentation. Further more, the fusion of the two memory module additionally achieve a promising result. Note that, we conduct late fusion for the two memory module, which is better than early fusion in our experiment.

\begin{figure}[ht]
    \centering
    \includegraphics[width=8cm]{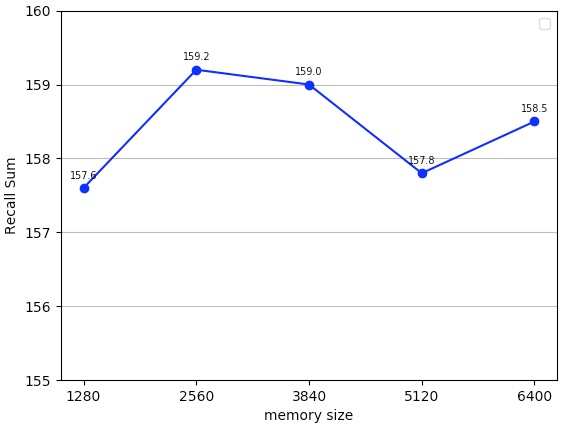}
    \caption{Evaluation of the proposed cross-modal memory module with different memory sizes.}
    \label{Figure,memory_size}
\end{figure}

\textbf{Impact of momentum encoder.} As mentioned in Section \ref{sec,cross-modal memory module}, to avoid the fast evolving of the embedding in memory bank, we conduct an extra momentum encoder for the cross-modal memory module. As a result, during the encoding, there are two kinds of embedding: embedding with momentum encoder and embedding without momentum encoder as illustrated in Figure \ref{Figure,model}. We compare the retrieval performance of best model selected by these two kinds embedding as shown in the second row and third row of Table \ref{table,msrvtt_ab}. The improvement demonstrates the effectiveness of momentum encoder.

\textbf{Impact of memory size.} To evaluate the impact of memory size $K$ of the proposed cross-modal memory module, we remove the text center memory for precise comparison. We set $K$ to be an integer multiple of the batch size. The results are shown in Figure \ref{Figure,memory_size}. From Figure \ref{Figure,memory_size}, we can see that the retrieval performance is not sensitive to $K$, while $K=2560$ yields the best performance.

\subsection{Qualitative Results}

To better understand the contribution of our proposed MEEL, we compare some visualization of retrieval results on MSR-VTT testing split, which are respectively retrieved by the models trained with and without MEEL. Figure \ref{Figure,result_t2i} illustrates the text-to-video retrieval examples. In the three visualization examples, we can observe that: giving a query sentence, DualEncoding trained without our MEEL can be always confused by some similar video, while the model trained with MEEL can effectively find the correct video. This may contributes to the global negative mining, which makes the model have the ability to distinguish between similar videos that will make confusion. We provide the video-to-text retrieval examples in Figure \ref{Figure,result_i2t} as well, where similar phenomena can be observed in the three giving example. Specifically, the model trained with MEEL can give more correct sentences with a high rank, because we consider to bridge the difference between these sentences. Thus the effectiveness of our proposed MEEL is demonstrated through these examples.

\section{CONCLUSION}

\begin{figure*}[htbp]
    \centering
    \includegraphics[width=13cm,height=9cm]{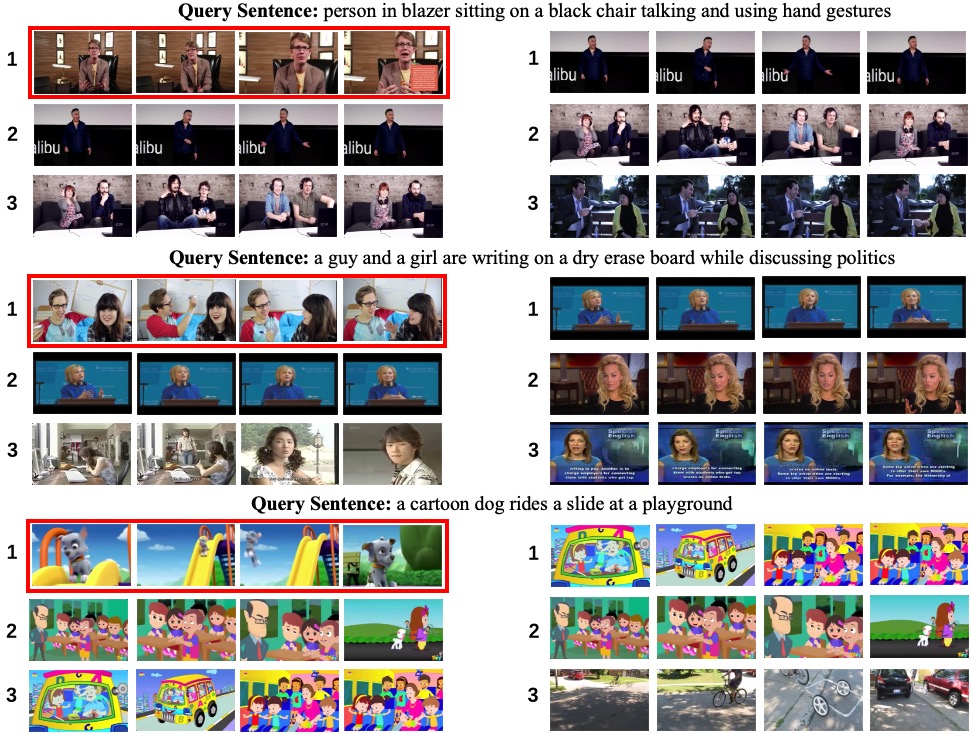}
    \caption{Top 3 text-to-video retrieval examples on MSR-VTT, where videos in the left column are retrieved by DualEncoding with MEEL and videos in the right column are retrieved without MEEL  (red boxes are correct).}
    \label{Figure,result_t2i}
\end{figure*}
\begin{figure*}
    \centering
    \includegraphics[width=14cm]{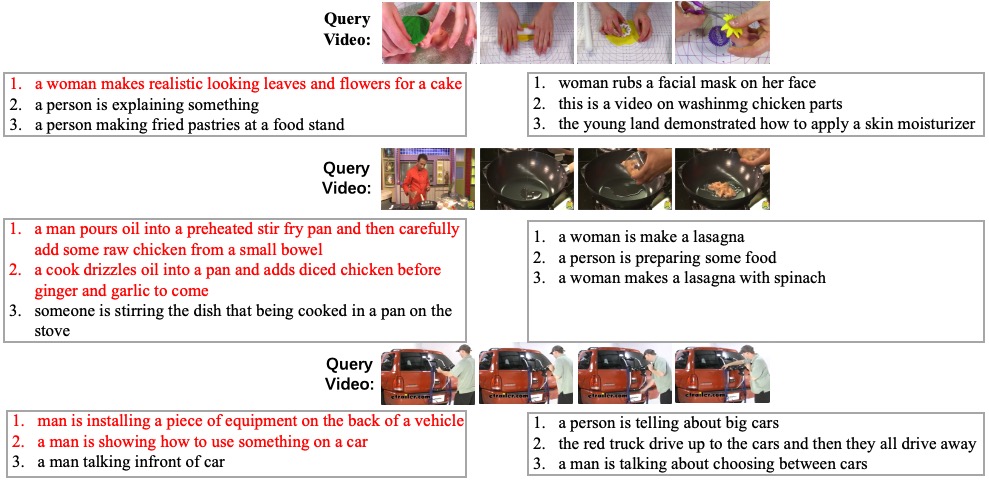}
    \caption{Top 3 video-to-text retrieval examples on MSR-VTT, where texts in the left column are retrieved by DualEncoding with MEEL and texts in the right column are retrieved without MEEL (red sentences are correct).}
    \label{Figure,result_i2t}
\end{figure*}
In this paper, we proposed a novel memory enhanced embedding learning for video-text retrieval, which considers two key properties existing in this task: global negative mining and the one video paired with multiple texts. Specifically, to unify these two problems in a framework, we fully incorporate the memory module into two types: cross-modal memory module and text center memory module. The cross-modal memory that stores more previous embeddings are constructed for global negative mining, which consists of video-to-text memory and text-to-video memory. The text center memory that stores the center information of multiple texts belonging to one video are constructed to bridge between these texts.  We conduct extensive experiments on two datasets: MSR-VTT and VATEX. The expermental results have demonstrated the effectiveness of our proposed method.


\bibliographystyle{ACM-Reference-Format}
\bibliography{sample-base}

\end{document}